\documentclass[sigplan,screen,9pt]{acmart}
\AtBeginDocument{%
  \providecommand\BibTeX{{%
    \normalfont B\kern-0.5em{\scshape i\kern-0.25em b}\kern-0.8em\TeX}}}

\usepackage{multirow} 

\usepackage[linesnumbered,lined,ruled]{algorithm2e}
 
\usepackage{arydshln} 
\usepackage{graphicx}
\usepackage{subcaption}
\usepackage{paralist}

\copyrightyear{2024} 
\acmYear{2024} 
\setcopyright{acmlicensed}\acmConference[DAC '24]{61st ACM/IEEE Design
Automation Conference}{June 23--27, 2024}{San Francisco, CA, USA}
\acmBooktitle{61st ACM/IEEE Design Automation Conference (DAC '24), June
23--27, 2024, San Francisco, CA, USA}
\acmDOI{10.1145/3649329.3655917}
\acmISBN{979-8-4007-0601-1/24/06}

\begin{document}

\title{
AdaptiveFL:
Adaptive Heterogeneous Federated Learning for  Resource-Constrained AIoT Systems
}

\author{Chentao Jia}
\affiliation{%
  \institution{East China Normal University}
  \streetaddress{200062}
  \city{Shanghai}
  \country{China}
}

\author{Ming Hu}
\affiliation{%
  \institution{Nanyang Technological University}
  \city{Singapore}
  \country{Singapore}
}
\authornote{Corresponding authors}
\email{hu.ming.work@gmail.com}

\author{Zekai Chen}
\affiliation{%
  \institution{East China Normal University}
  \streetaddress{200062}
  \city{Shanghai}
  \country{China}
}

\author{Yanxin Yang}
\affiliation{%
  \institution{East China Normal University}
  \streetaddress{200062}
  \city{Shanghai}
  \country{China}
}

\author{Xiaofei Xie}
\affiliation{%
  \institution{Singapore Management University}
  \city{Singapore}
  \country{Singapore}
}

\author{Yang Liu}
\affiliation{%
  \institution{Nanyang Technological University}
  \city{Singapore}
  \country{Singapore}
}

\author{Mingsong Chen}
\affiliation{%
  \institution{East China Normal University}
  \streetaddress{200062}
  \city{Shanghai}
  \country{China}
}
\authornotemark[1]
\email{mschen@sei.ecnu.edu.cn}

\renewcommand{\shortauthors}{Chentao Jia, et al.}

\begin{abstract}
Although Federated Learning (FL) is promising to enable collaborative learning among Artificial Intelligence of Things (AIoT) devices, it suffers from the problem of low classification performance due to various heterogeneity factors (e.g., computing capacity, memory size) of devices and uncertain operating environments. 
To address these issues, this paper introduces an effective FL approach named {\it AdaptiveFL} based on a novel fine-grained width-wise model pruning mechanism, which can generate various heterogeneous local models for heterogeneous AIoT devices. 
By using our proposed reinforcement learning-based device selection strategy, 
AdaptiveFL can adaptively dispatch suitable heterogeneous models to corresponding AIoT devices based on their available resources for local training.  
Experimental results show that, compared to state-of-the-art methods, AdaptiveFL can achieve up to 8.94\% inference improvements for both IID and non-IID scenarios.

\end{abstract}

\maketitle

\section{Introduction}


Although Federated Learning (FL)~\cite{FedAvg} has been increasingly 
studied in Artificial Intelligence of Things (AIoT) design~\cite{hu2023aiotml,zhang_tacd_2021,hu2023gitfl} to enable 
 knowledge sharing without compromising data privacy among devices, 
it suffers from the problems of large-scale deployment and low inference accuracy. 
This is mainly because most existing FL methods assume that the models on device are homogeneous. When dealing with an AIoT system involving 
devices with various heterogeneous hardware resource constraints (e.g., computing capability, memory size), the overall inference performance of existing FL approaches is often greatly limited, especially when the data on devices are non-IID (Independent and Identically Distributed).
To address this problem, various heterogeneous FL methods~\cite{heterofl, depthfl, ScaleFL, compelete1, compelete2} have been proposed, which can be 
classified into two categories, i.e., {\it completely heterogeneous} and {\it partially heterogeneous} methods.
The completely heterogeneous approaches~\cite{compelete1, compelete2} rely on both device models with different structures for local training and knowledge distillation technologies to facilitate knowledge sharing among these models. 
As an alternative, the partially heterogeneous methods~\cite{heterofl, depthfl, ScaleFL} adopt hypernetworks as full global models, which can be used to generate various heterogeneous device models to enable model aggregation based on specific model pruning mechanisms.

Although the state-of-the-art heterogeneous FL methods 
can improve the overall inference performance of devices, 
most of them cannot be directly applied to  AIoT systems. As an example
of completely heterogeneous FL, the need for extra 
high-quality datasets may violate the data privacy requirement. 
Meanwhile, due to the Cannikin Law, the learning capabilities of completely heterogeneous FL methods are 
determined by small models, which are usually 
hosted by weak devices with fewer data samples. 
Similarly, for partially heterogeneous FL  methods, the coarse-grained 
pruning on hypernetworks may weaken the learning capabilities
of the model. Things become even worse when such FL-based AIoT systems are deployed in uncertain environments~\cite{hu2020quantitative} with dynamically changing 
available resources, since the assignment of 
improperly pruned models to devices will inevitably result in 
insufficient learning of local models, thus hampering the overall inference capability of AIoT systems. 
\textit{Therefore, how to wisely and adaptively assign properly pruned heterogeneous models to devices in order to maximize the overall inference performance of all the involved heterogeneous models in resource-constrained scenarios is becoming a major challenge in FL-based AIoT systems.}

Intuitively, to alleviate the insufficient training of local models, heterogeneous models should share more key generalized parameters.
For a Deep Neutral Network (DNN), the number of parameters of shallow layers is typically smaller than those of deep layers.
According to the observation in~\cite{li2016pruning}, 
pruning shallow layers rather than deep layers can result in greater performance degradation.
In other words, if the pruning happens at deep layers of DNN, the 
inference performance degradation of DNN is negligible, while the size of the models can be significantly reduced.
Inspired by this fact,  this paper presents an effective heterogeneous FL approach named AdaptiveFL, which uses a novel fine-grained width-wise model pruning mechanism to generate heterogeneous models for local training. 
In AdaptiveFL, devices can adaptively prune received models to accommodate their available resources. Since AdaptiveFL does not prune any entire layer, the pruned models can be trained directly on devices without additional parameters or adapters.
To avoid exposing 
 device status to the cloud server, 
 AdaptiveFL adopts a Reinforcement Learning (RL)-based device selection strategy, which can select the most suitable devices to train models with specific sizes based on the historical size information of trained models. In this way, the communication
waste caused by dispatching mismatched models can be drastically reduced. 
 This paper makes the following three major contributions: 
\begin{itemize}
    \item We propose a fine-grained width-wise pruning mechanism to wisely and adaptively generate heterogeneous models in resource-constrained scenarios.

    
    \item We present a novel RL-based device selection strategy to select devices with suitable hardware resources for the given heterogeneous models, which can reduce the communication waste caused by dispatching mismatched large models.
    
    \item We perform extensive simulation and real test-bed experiments to evaluate the performance of AdaptiveFL. 

\end{itemize}

\section{Background and Related Work}
\textbf{Preliminaries to FL.}
Generally, an FL  system consists of one cloud server and multiple dispersed clients. In each round, the cloud server will first send the global model to selected devices. After receiving the model, the devices conduct local training and upload the parameters of the model to the cloud server. Finally, the cloud server aggregates the received parameters to update the original global model. So far, almost all FL methods aggregate local models based on FedAvg\cite{FedAvg} defined as follows:
\begin{equation}
\begin{split}
    \min _w F(w)=\frac{1}{K} \sum_{k=1}^K f_k(w), \text { s.t., } f_k(w)=\frac{1}{\left|d_k\right|} \sum_{i=1}^{\left|d_k\right|} \ell\left(w,\left\langle x_i, y_i\right\rangle\right),
    \nonumber
\end{split}
\end{equation}
where $K$ is the total number of clients, $\left|d_k\right|$ is the number of data samples hosted by the $k^{th}$ client, $\ell$ denotes loss function (e.g., cross-entropy loss), $x_i$ denotes a sample, and $y_i$ is the label of $x_i$.

\textbf{Model Heterogeneous FL.}
Model heterogeneous FL has a natural advantage in solving the problem of system heterogeneity, where submodels of different sizes can better fit heterogeneous clients.
Relevant prior work includes studies of width-wise pruning, depth-wise pruning, and two-dimensional scaling.
For width-wise pruning, Diao \textit{et al.}~\cite{heterofl} proposed HeteroFL, which prunes model architectures for clients with variant widths and conducted parameter-averaging over heterogeneous models.
For depth-wise pruning, Kim \textit{et al.}~\cite{depthfl} proposed DepthFL, which obtains local models of different depths by pruning the deepest layers of the global model.
Recently, Ilhan \textit{et al.}~\cite{ScaleFL} proposed a two-dimensional pruning approach called ScaleFL, which utilizes self-distillation to transfer the knowledge among submodels. 
However, existing approaches seldom consider the resource uncertainties associated with devices in real-world environments. Most of them employ a coarse-grained way for model pruning. In addition, resource information is the key to dispatch the appropriate model for each client in their approach, yet in practical applications, obtaining accurate resource information for devices can be difficult.

To the best of our knowledge, AdaptiveFL is the first 
resource-adaptive FL framework for heterogeneous AIoT
devices without collecting their resource information. Since AdaptiveFL adopts a fine-grained width-wise model pruning mechanism together with our proposed RL-based device selection strategy, it can be easily integrated into large-scale AIoT systems to maximize knowledge sharing among devices.


\section{Our Approach}

\begin{figure*}[h]
  \centering
  \includegraphics[width =0.8\linewidth]{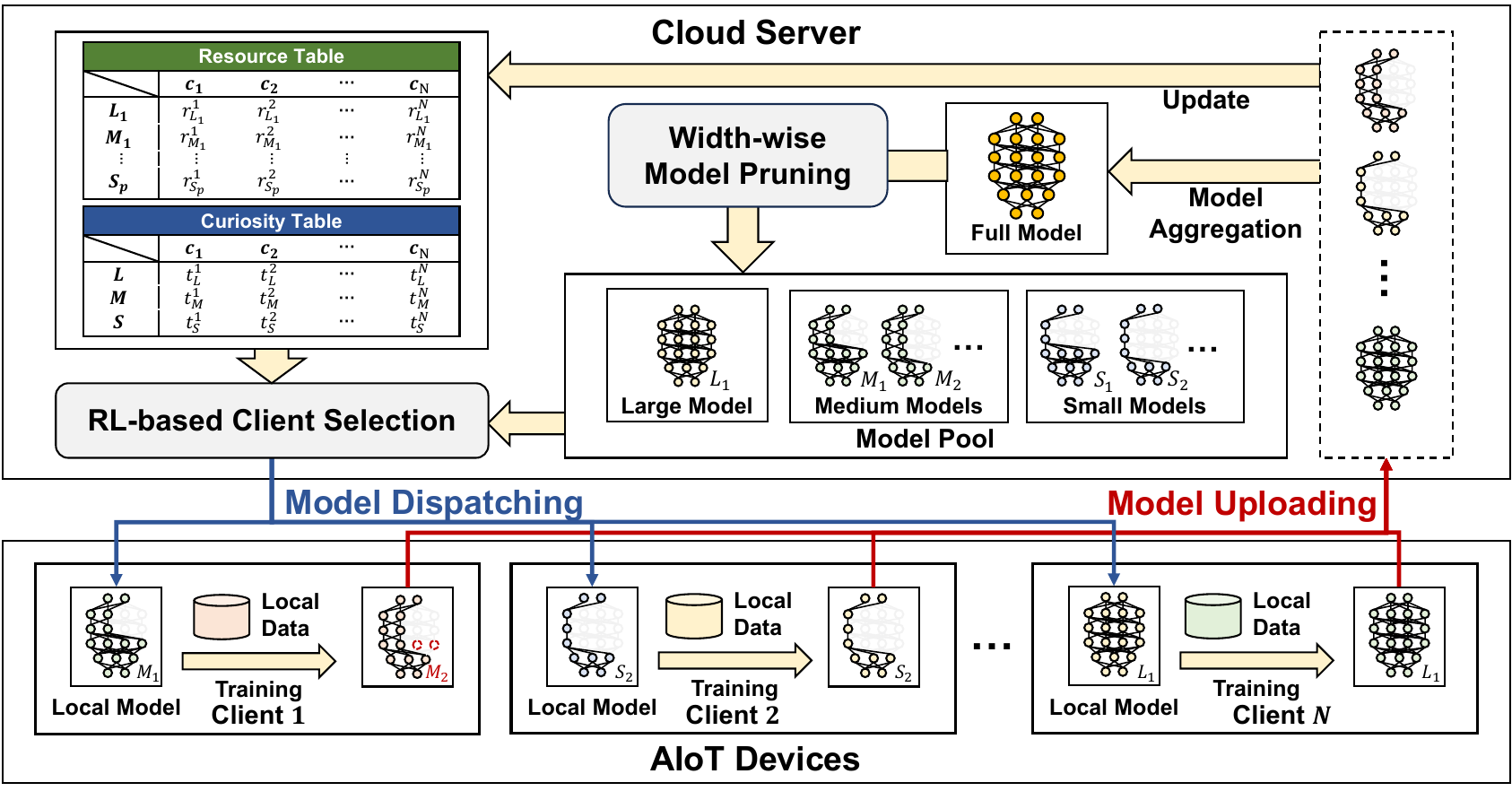} 
  \caption{Framework and workflow of AdaptiveFL.}
  \label{fig:framework}
  \vspace{-0.1in}
\end{figure*}

Figure \ref{fig:framework} presents the framework and workflow of AdaptiveFL. 
As shown in the figure, the cloud server performs three key stages, i.e., model pruning, RL-based device selection, and model aggregation.
In the model pruning stage, the cloud server prunes the entire global model into multiple heterogeneous models, which will be dispatched to devices for local training.
In the RL-based device selection stage, the cloud server selects a best-fit device for each heterogeneous model based on the curiosity table and the resource table. 
In the model aggregation stage, the cloud server aggregates the weights of uploaded models and updates the global model.

In specific, each FL training round of AdaptiveFL includes six key steps as follows:
\begin{itemize}
    \item \textbf{Step 1: Model Pruning}. The cloud server generates multiple heterogeneous models based on the full global model by using the \textit{fine-grained width-wise model pruning} mechanism and stores the generated models to the model pool;
    \item \textbf{Step 2: Model Selection}. The cloud server randomly selects a list of generated heterogeneous models from the model pool as dispatched models for local training;
    \item \textbf{Step 3: Client Selection}. The cloud server selects a client for each dispatching model by using our RL-based selection strategy and dispatches the model to its selected client;
    \item \textbf{Step 4: Local Training}. AIoT devices adaptively prune the received model according to their local available resources and train the model on their local raw data;
    \item \textbf{Step 5: Model Uploading}. Devices upload the trained model to the cloud server;
    \item \textbf{Step 6: Model Aggregation}. The cloud server generates a new global model by aggregating the corresponding parameters of all the uploaded models.
\end{itemize}

\subsection{Implementation of AdaptiveFL}\label{subsection: Implementation of AdaptiveFL}

Algorithm \ref{algorithm: AdaptiveFL} details the implementation of AdaptiveFL.  
Before FL training, Lines 1-2 initialize the curiosity table $T_c$ and the resource table $T_r$.
Lines 3-29 present the details of FL training for each round.
Line 4 splits the global model $M$ into submodels in different size levels (i.e., small, medium and large) and stores them in model pool $R = \{ m_{S_p},m_{S_{p-1}},\ldots, m_{M_2},m_{M_1},m_{L_1}\}$, where hyperparameters $p$ is the number of submodels in each level except $L$ level, and it should be noted that the large model $m_{L_1}$ is unpruned which is equivalent to the global model.
Lines 6-27 present the FL training process of the models waiting for training, where the loop ``for'' is parallel.
In Line 7, the function \textit{RandomSel(.)} is to randomly select a model $m_i$ from the model pool $R$.
In Line 8, the function \textit{ClientSel(.)} is to select a suitable client $c_i$ for model $m_i$ from client set $C$ based on RL-table $T_c$ and $T_r$.
In Line 9, the function \textit{LocalTrain(.)} is to dispatch the model $m_i$ to the selected client $c_i$ for local training, and return the trained model $m_i^{\prime}$ with the local data size $\left|d_{c_i}\right|$ back to the server. 
Line 10 stores $m_i^{\prime}$ and $\left|d_{c_i}\right|$ to array $ML_{back}$ and array $Len$, respectively, which are used for aggregation later.
In addition, Lines 12-13 and Lines 14-26 present the updating process of $T_c$ and $T_r$, respectively.
In Lines 12-13, we updated the selection times for the level of the send and back models in $T_c$, respectively, where $type\left(m_i\right)$ means the level of model $m_i$, e.g., $type\left(m_{S_p}\right)$ return the size level $S$.
As for the update of $T_r$, we consider the following two cases: i) In Lines 15-18, since no pruning is done locally at the client $c_i$, which means that the resource capacity $\Gamma_{c_i} \geq \operatorname{size}\left(m_i=m_i^{\prime}\right)$, so we perform an increment operation for the training score in the table whose model size is larger than $m_i$; ii) In Lines 20-25, it shows that $\operatorname{size}\left(m_i^{\prime}\right) \leq \Gamma_{c_i} \leq \operatorname{size}\left(\hat{m_i^{\prime}}\right)$, $\hat{m_i^{\prime}}$ here is the nearest greater model with $m_i^{\prime}$ in $R$. Thus, we use a penalty term $\tau$ to reduce the training score of the heterogenous model that is larger than $\hat{m_i^{\prime}}$, while increasing the training score of the model $m_i^{\prime}$.

\begin{algorithm}[htp]
        \caption{Implementation of AdaptiveFL}
        \label{algorithm: AdaptiveFL}
        \KwIn{i) $T$, training rounds; ii) $C$, client set; iii) $K$, the number of clients selected each round; iv) $p$, the number of model in each level.}
        $T_c[i][j] \leftarrow 1 $ for $i \in[1,3], j \in[1,|C|]$\\
        $T_r[i][j] \leftarrow 1 $ for $i \in[1,2p+1], j \in[1,|C|]$\\
        \For{epoch $ E = 1,\ldots,T$}{
             $ R = \{ m_{S_p},m_{S_{p-1}},\ldots, m_{M_2},m_{M_1},m_{L_1}\} \leftarrow \operatorname{Split}\left(M\right)$\\
            /*parallel for*/\\
            \For{$i=1, \ldots, K$ }{
                $m_i \leftarrow \operatorname{RandomSel}\left(R\right)$\\
            
                $c_i \leftarrow \operatorname{ClientSel} \left(m_i, T_c, T_r, C\right)$  //RL-based Client Selection\\
    
                $\left(m_i^{\prime}, \left|d_{c_i}\right|\right)\leftarrow  \operatorname{LocalTrain}\left(c_i, m_i\right)$ // Local Training\\
                $ML_{back}[i] \leftarrow m_i^{\prime}$,  $Len[i]\leftarrow \left|d_{c_i}\right|$\\
                /* Update RL Table */\\

                $T_c[type\left(m_i\right)][c_i] \leftarrow T_c[type\left(m_i\right)][c_i]+1$\\
                $T_c[type\left(m_i^{\prime}\right)][c_i] \leftarrow T_c[type\left(m_i^{\prime}\right)][c_i]+1$\\
                \eIf{$m_i == m_i^{\prime}$}{
                    \For{$t = m_i,\ldots,m_{L_1}$}{
                        $T_r[t][c_i] \leftarrow T_r[t][c_i]+1$
                    }
                    $T_r[m_{L_1}][c_i] \leftarrow T_r[m_{L_1}][c_i]+p-1$
                }
                {
                    $T_r[m_i^{\prime}][c_i] \leftarrow T_r[m_i^{\prime}][c_i]+p$\\
                    $\tau\leftarrow 0$\\
                    \For{$t = m_i^{\prime},\ldots,m_{L_1}$}{
                        $T_r[t][c_i] \leftarrow \operatorname{max}\left(T_r[t][c_i]-\tau, 0\right)$\\
                        $\tau\leftarrow \tau + 1$
                    }
                }
            }
            
            $M \leftarrow \operatorname{Aggregate}(m_{L_1}, ML_{back}, Len)$ \\
        }
\end{algorithm}

\subsection{Fine-Grained Width-Wise Model Pruning Mechanism}\label{subsection: Fine-grained Width-wise Model Pruning}
To enable devices to prune models according to their available resources adaptively, we adopt a width-wise pruning mechanism where 
the pruned model can be trained directly without additional adapters or parameters.
Inspired by the observations in \cite{li2016pruning}, 
we prefer to prune the parameters of deep layers, which enables large models trained by insufficient data to achieve higher performance.
Specifically, our fine-grained width-wise model pruning mechanism is controlled by two hyperparameters, i.e., the width pruning ratio $r_w$ and the index of the starting pruning layer $I$, respectively, where adjusting $r_w$ can significantly change the model size while adjusting $I$ can fine-tune the model size.

\textbf{Width-Wise Model Pruning ($r_w$).}
To generate multiple models of different sizes, the cloud server prunes partial kernels in each layer of the model, where the number of kernels pruned is determined by the width pruning ratio $r_w \in (0,1]$.
Specifically, we assume that $W_g$ is the parameter of the global model $M_g$, $d_k$ and $n_k$ denote the output and input channel size of the $k^{th}$ hidden layer of $M_g$, respectively.
Then the parameters of the $k^{th}$ hidden layer can be denoted as $W_g^{k} \in \mathbb{R}^{d_k \times n_k}$.
With a width-wise pruning ratio $r_w$, the pruned weights of the $k^{th}$ hidden layer can be presented as $W_{r_w}^k = W_g^k[:d_k\times r_w][:n_k \times r_w]$.

\textbf{Layer-Wise Model Adjustment ($I$).} 
To address performance fluctuations caused by uncertainty, our pruning mechanism supports fine-tuning the model size by adjusting the index of the starting pruning layer $I$.
Note that to ensure that heterogeneous models share shallow layers, the index of the starting pruning layer must be set larger than the specific threshold $\tau$.
Specifically, assume that $I \geq \tau$, the weights of the $k^{th}$ layer can be presented as $W_{r_w}^k = W_g^k[:d_k][:n_k]$ when $k \leq I$, and which can be presented as $W_{r_w}^k = W_g^k[:d_k\times r_w][:n_k \times r_w]$ when $k > I$.


\textbf{Available Resource-Aware Pruning.}
To prevent failed training caused by limited resources, our pruning mechanism supports each device in pruning the received model adaptively according to its available resources.
Specifically, assume that the available resource capacity of the device is $\Gamma$, the weight of the received model is $W$ and $I \geq \tau$, and the width-wise pruning ratio $r_w$ and the index of the starting pruning layer $I$ can be determined as follows:
\begin{equation}\label{eq: model pruning config}
    \begin{split}
        \mathop{\arg\max}\limits_{r_w,I} size(prune(W ; r_w,I)),\\
    \text{s.t.,}\ size(prune(W ; r_w,I)) \leq \Gamma\ \text{and}\ I \geq \tau,
    \nonumber 
    \end{split}
\end{equation}

\subsection{RL-based Client Selection}\label{subsection: RL-based Client Selection}
Due to uncertainty and privacy concerns, the cloud server cannot obtain available resource information for AIoT devices.
To avoid communication waste caused by dispatching unsuitable models, we propose an RL-based device selection strategy.
By utilizing the information of historical dispatching and the corresponding received model $\langle m_i, m_i^{\prime}\rangle$ of clients, RL can learn the information about the available resources of each device. 
Based on the learned information, RL can learn a strategy to select suitable devices for each heterogeneous model wisely.




\textbf{Problem Definition.}
In our approach, the client selection process can be regarded as a Markov Decision Process~\cite{hu_rtss2021}, which can be presented as a four-tuple $MDP=\langle\mathcal{S}, \mathcal{A}, \mathcal{F}, \mathcal{R}\rangle$ as follows:
\begin{itemize}
    \item $\mathcal{S}$ is a set of states. We use a vector $s_t=\left\langle D_t, S, C_t, T_c, T_r\right\rangle$
to denote the state of AdaptiveFL, where $D_t$ denotes the set
of submodels that wait for dispatching, $S$ is the list of the size information of all submodels in the model pool, $C_t$ indicates the set of clients involved, 
$T_c$ and $T_r$ are the curiosity table and the resource table, respectively.
    \item $\mathcal{A}$ is a set of actions. At the state of $s_t=\left\langle D_t, S, C_t, T_c, T_r\right\rangle$,
the action $a_t$ aims to select a suitable client $c_i \in C_t$ for the candidate model $m_i \in D_t$.
    \item $\mathcal{F}$ is a set of transitions. It records the transition $s_t \stackrel{a_t}{\longrightarrow} s_{t+1}$ with the action $a_t$.
    \item $\mathcal{R}$ is the reward function. We combine the values in the resource table $T_r$ and the curiosity table $T_c$ of each client as the reward to guide the selection on this round.
\end{itemize}

\textbf{Resource- and Curiosity-Driven Client Selection.}
Since there is an implicit connection between the model size with the resource budget, the model returned by the client can be used to determine the available resource range of the device.
Specifically, AdaptiveFL uses a client resource table $T_r$ to record the historical training score for each heterogeneous model on each client, where a higher score indicates that the client has a higher success rate in training the corresponding model.
The resource reward for client $c$ on submodel $m_i$ is measured as follows:
\begin{equation}
    \begin{split}
        R_s(m_i, c)=  \frac{\sum_{k=T_p, T=\operatorname{type}\left(m_i\right),k \in R}^{T_1}\sum_{t = k}^{L_1} T_r[m_t][c]}{p \times \sum_{k = S_p, k \in R}^{L_1} T_r[m_k][c]}.
        \nonumber
    \end{split}
\end{equation}

To balance the training times of the same model level on different clients, we utilize curiosity-driven exploration~\cite{hu2023accelerating, hu2023gitfl} as one of the reward evaluation strategies, while the client who is selected fewer times on a size level of the model will get higher curiosity rewards. AdaptiveFL uses the curiosity table $T_c$ to record the selection times of each client on a type of model, and performs Model-based Interval Estimation with Exploration Bonuses (MBIE-EB)~\cite{bellemare2016unifying} to calculate the curiosity reward as follows: 
\begin{equation}
    \begin{split}
        R_c(m_i, c)=\frac{1}{\sqrt{T_c[type\left(m_i\right)][c]}},
        \nonumber
    \end{split}
\end{equation}
where $T_c[type\left(m_i\right)][c]$ indicates the total selection number of model type $type\left(m_i\right)$ on client $c$. 
To avoid the higher success rate of the large client leading to the lower probability of other clients being selected, we set the upper success rate of 50\%, and the selection of clients whose success rate is beyond 50\% will be determined by the curiosity reward. 
Consequently, the final reward for each client on the model $m_i$ is
calculated by combining resource reward $R_s$ and curiosity reward $R_c$ as follows:
\begin{equation}
    \begin{split}
        R(m_i, c)=\min \left(0.5, R_s(m_i, c)\right) \times R_c(m_i, c).
        \nonumber
    \end{split}
\end{equation}

In conclusion, based on the final reward, the probability that the client $c$ is selected for model $m_i$ is:
\begin{equation}
    \begin{split}
        P(m_i, c)=\frac{R(m_i, c)}{\sum_{j=1}^{|C|} R(m_i, j)}.
        \nonumber
    \end{split}
\end{equation}

\subsection{Heterogenous Model Aggregation.}\label{subsection: Heterogenous Model Aggregation.}

\begin{algorithm}[h]
        \caption{Heterogenous Model Aggregation}
        \label{algorithm: Aggregation}
        \KwIn{i) $\theta=\{l^\theta_1, ..., l^\theta_N\}$, global model weights; ii) $\{\theta_{c}\}_{c\in S}$, set of local model weights; iii) $\{\left|d_c\right|\}_{c\in S}$, set of local data size }
        \KwOut{$\theta^{\prime}$, aggregated global model weights}
        $\theta^{\prime} \leftarrow Zero(\theta)$\\
        \For{$k$ in  $1,..,N$}{
            
            $L_w\leftarrow Zero(len(l^{\theta^\prime}_k))$\\
            \For{ $c$ in $S$}{
                \For{ $i$ in $1, ..., len(l^{\theta_c}_k)$}{
                     $l^{\theta^\prime}_k[i] \leftarrow l^{\theta^\prime}_k[i] + l^{\theta_c}_k[i] \times |d_c|$\\
                     $L_w[i]\leftarrow L_w[i] + |d_c|$
                }
            }

            \For{ $j$ in $1, ..., len(l^{\theta^\prime}_k)$}{
                \eIf{$L_w[j] > 0$}{
                     $l^{\theta^\prime}_k[j] \leftarrow \frac{l^{\theta^\prime}_k[j]}{L_w[j]}$\\
                }{
                     $l^{\theta^\prime}_k[j] \leftarrow l^{\theta}_k[j]$
                }
            }
        }
        \textbf{return} $\theta^{\prime}$
\end{algorithm}


In our model pruning mechanism, since all the submodels are pruned based on the same full global model, the cloud server can update the global model by aggregating all the received heterogenous submodels according to the corresponding index of their parameters in the full model.
Algorithm \ref{algorithm: Aggregation} details the aggregation process of our approach. Line 1 is the initialization of the process. Lines 2-17 update the parameters of each layer in the model. Line 3 initializes the variable $L_w$, which is used to count the total number of the training data size for each parameter. In Lines 4-8, the model parameters of each client are added with weights, which is the size of local data. For each client, the uploaded model often lacks some parameters compared to the complete model. Lines 10-16 take the average of the updated parameters. Note that if some parameters are not included in any uploaded model, they will keep their original values unchanged, which is shown in Line 14.


\section{Performance Evaluation}
To evaluate the performance of AdaptiveFL, we implemented it using PyTorch. For a fair comparison, we adopted the same SGD optimizer with a learning rate of 0.01 and a momentum of 0.5 for all the investigated FL methods. For local training, we set the batch size to 50 and the local epoch to 5. All the experiments were conducted on a Ubuntu workstation with one Intel i9 13900k CPU, 64GB memory, and one NVIDIA RTX 4090 GPU. 

\subsection{Experimental Settings}
\textbf{Data Settings.} 
We conducted experiments on three well-known datasets, i.e., CIFAR-10, CIFAR-100~\cite{CIFAR}, and FEMNIST~\cite{LEAF}. For both CIFAR-10 and CIFAR-100, we assumed that there were 100 clients participating in FL. For FEMNIST, there were 180 clients involved. In each round, 10\% of the clients will be selected for local training. We considered both IID and non-IID scenarios for CIFAR-10 and CIFAR-100, where we adopted the Dirichlet distribution
to control the data heterogeneity. Here, the smaller the coefficient $\alpha$, the higher the heterogeneity of the data. Note that  FEMNIST is naturally non-IID distributed. 

\begin{table}[h]
\caption{Split settings for VGG16 ($p=3$).}
\label{table: splitting settings of vgg model}
\footnotesize
\addtolength{\tabcolsep}{3pt}
\begin{tabular}{c|cc|ccc}
\hline
VGG16 & \multicolumn{2}{c|}{Pruning Configuration} & \multicolumn{3}{c}{Model Size} \\ \hline
\textbf{Level} & $r_w$                    & $I$   & \#PARAMS  & \#FLOPS  & \textit{ratio}  \\ \hline
$L_1$     & 1.00                     & N/A   & \textbf{33.65M}    & \textbf{333.22M}  & \textbf{1.00}    \\
\hdashline
$M_1$    & \multirow{3}{*}{0.66}    & 8     & \textbf{16.81M}    & \textbf{272.17M}  & \textbf{0.50}    \\
$M_2$     &                          & 6     & 15.41M    & 239.95M  & 0.46    \\
$M_3$     &                          & 4     & 14.84M    & 203.41M  & 0.44    \\
\hdashline
$S_1$     & \multirow{3}{*}{0.40}    & 8     &\textbf{ 8.39M}     & \textbf{239.00M}  &\textbf{ 0.25}    \\
$S_2$     &                          & 6     & 6.48M     & 191.31M  & 0.19    \\
$S_3$     &                          & 4     & 5.67M     & 139.07M  & 0.17    \\ \hline
\end{tabular}
\end{table}

\begin{table*}[h]
\caption{Test accuracy (\%) comparison of avg/full models (the best and second-best results are in \textbf{bold} and \underline{underlined}, respectively). }
\addtolength{\tabcolsep}{1pt}
\label{table: Local and global model accuracy comparison}
\footnotesize
\begin{tabular}{cl|llllll|llllll|ll}
\hline
\multirow{3}{*}{Model}    & \multicolumn{1}{c|}{\multirow{3}{*}{Algorithm}} & \multicolumn{6}{c|}{CIFAR-10}                                                                                            & \multicolumn{6}{c|}{CIFAR-100}                                                                                           & \multicolumn{2}{c}{FEMNIST}     \\ \cline{3-16} 
                          & \multicolumn{1}{c|}{}                           & \multicolumn{2}{c}{IID}                & \multicolumn{2}{c}{$\alpha$ = 0.6}            & \multicolumn{2}{c|}{$\alpha$ = 0.3}           & \multicolumn{2}{c}{IID}                & \multicolumn{2}{c}{$\alpha$ = 0.6}            & \multicolumn{2}{c|}{$\alpha$ = 0.3}           & \multicolumn{2}{c}{-}           \\ \cline{3-16} 
                          & \multicolumn{1}{c|}{}                           & avg                 & full         & avg                 & full         & avg                 & full         & avg                 & full         & avg                 & full         & avg                 & full         & avg          & full         \\ \hline
\multirow{5}{*}{VGG16}    & All-Large~\cite{FedAvg}                                          & \multicolumn{1}{c}{-} & \underline{79.76}          & \multicolumn{1}{c}{-} & \underline{77.29}          & \multicolumn{1}{c}{-} & \underline{74.95}          & \multicolumn{1}{c}{-} & \underline{40.71}          & \multicolumn{1}{c}{-} & \textbf{41.13} & \multicolumn{1}{c}{-} & \underline{40.34}          & \multicolumn{1}{c}{-}   & \underline{85.21}          \\
                          & Decoupled~\cite{FedAvg}                                       & 75.02                 & 69.80           & 72.95                 & 67.58          & 69.11                 & 62.91          & \underline{33.66}                 & 26.67          & \underline{33.37}                 & 26.53          & \underline{32.86}                 & 26.54          & \underline{78.45}          & 70.13          \\
                          & HeteroFL~\cite{heterofl}                                        & 77.98                 & 74.96          & 75.18                 & 72.69          & 71.18                 & 67.59          & 32.22                 & 28.13          & 32.92                 & 28.82          & 32.32                 & 28.68          & 77.69          & 71.75          \\
                          & ScaleFL~\cite{ScaleFL}                                         & \underline{79.94}                 & 78.12          & \underline{76.08}                 & 75.07          & \underline{71.71}                 & 70.42          & 31.86                 & 32.17          & 30.82                 & 30.57          & 28.36                 & 29.61          & 71.58          & 67.36          \\
                          & \textbf{AdaptiveFL}                                         & \textbf{82.97}        & \textbf{83.14} & \textbf{81.12}        & \textbf{81.31} & \textbf{78.85}        & \textbf{78.99} & \textbf{40.61}        & \textbf{40.93} & \textbf{37.87}        & \underline{38.88}          & \textbf{40.95}        & \textbf{41.17} & \textbf{87.38} & \textbf{88.13} \\ \hline
\multirow{5}{*}{ResNet18} & All-Large~\cite{FedAvg}                                            & \multicolumn{1}{c}{-}                 & 68.37          & \multicolumn{1}{c}{-}                 & 67.03         & \multicolumn{1}{c}{-}                 & 64.28          & \multicolumn{1}{c}{-}                 & 35.08          & \multicolumn{1}{c}{-}                 & 34.74         & \multicolumn{1}{c}{-}                 & 33.84          & \multicolumn{1}{c}{-}              & \underline{83.94}          \\
                          & Decoupled~\cite{FedAvg}                                         & 63.23                 & 55.56          & 59.21                 & 52.59          & 55.82                 & 49.65          & 24.58                 & 22.35          & 25.22                 & 20.14          & 24.06                 & 20.02          & 74.37          & 65.20           \\
                          & HeteroFL~\cite{heterofl}                                        & 70.44                 & 65.37          & 65.97                 & 60.33          & 60.32                 & 55.83          & 30.43                 & 27.74          & 30.23                 & 23.59          & 28.96                 & 23.04          & 77.50           & 69.35          \\
                          & ScaleFL~\cite{ScaleFL}                                         & \underline{76.34}                 & \underline{76.51}          & \underline{72.68}                 & \underline{72.91}          & \underline{67.26}                 & \underline{67.50}           & \underline{40.30}                  & \underline{40.46}          & \underline{38.91}       & \underline{37.86}   & \underline{36.82}                 & \underline{36.56}          & \underline{83.64}          & 83.79          \\
                          & \textbf{AdaptiveFL}                                         & \textbf{77.14}        & \textbf{77.20}  & \textbf{74.72}        & \textbf{74.89} & \textbf{70.61}        & \textbf{70.97} & \textbf{41.09}        & \textbf{41.15} & \textbf{39.14 }                & \textbf{39.56 }         & \textbf{39.15}        & \textbf{39.65} & \textbf{87.11} & \textbf{87.30}  \\ \hline
\end{tabular}

\end{table*}

\textbf{Device Heterogeneity Settings.} 
To simulate the heterogeneity of devices, we set up three types of clients (i.e., weak, medium, and strong clients) and three levels of models (i.e., small, medium, and large models), where weak devices can only accommodate weak models,  medium devices can train medium or small models, while strong devices can accommodate models of any type. For the following experiments, we set the proportion of weak, medium, and strong devices to 4: 3: 3 by default. 
To show the generality of our AdaptiveFL framework, we conducted experiments based on two widely used models (i.e., VGG16~\cite{VGG} and ResNet18~\cite{ResNet}), where  Table \ref{table: splitting settings of vgg model} shows the split settings of VGG16.


\subsection{Performance Comparison}
We compared  AdaptiveFL with four baseline methods, i.e., 
 All-Large~\cite{FedAvg},  Decoupled~\cite{FedAvg}, HeteroFL~\cite{heterofl}, and ScaleFL~\cite{ScaleFL}. For 
 All-Large, we trained the $L_1$ model with all clients under the classic
FedAvg~\cite{FedAvg}. For Decoupled, we trained 
separate models (i.e., $L_1$, $M_1$, $S_1$ models) for each level using the available data of affordable clients. 
For HeteroFL and ScaleFL, we created their corresponding submodels at different levels. 
Table \ref{table: Local and global model accuracy comparison} shows the comparison results, 
 where the notations  ``avg'' and ``full'' 
denote the average accuracy of submodels at different levels (i.e., $L_1$, $M_1$, $S_1$) and the accuracy of the global model, respectively.

\textbf{Global Model Performance.}
From Table \ref{table: Local and global model accuracy comparison}, we can find that Decoupled has the worst inference performance in all the cases, since its submodels are only aggregated with the models at the same levels. However, AdaptiveFL can achieve up to 2.95\% and 3.12\% better
 inference than the second-best methods for ResNet18 and VGG16, respectively. 
 Note that AdaptiveFL can achieve better results compared with All-Large, indicating that AdaptiveFL can improve the FL performance in non-resource scenarios.


\begin{figure}[h]  
  \centering
  \begin{subfigure}{0.4\linewidth}
    \centering
    \includegraphics[width=\linewidth]{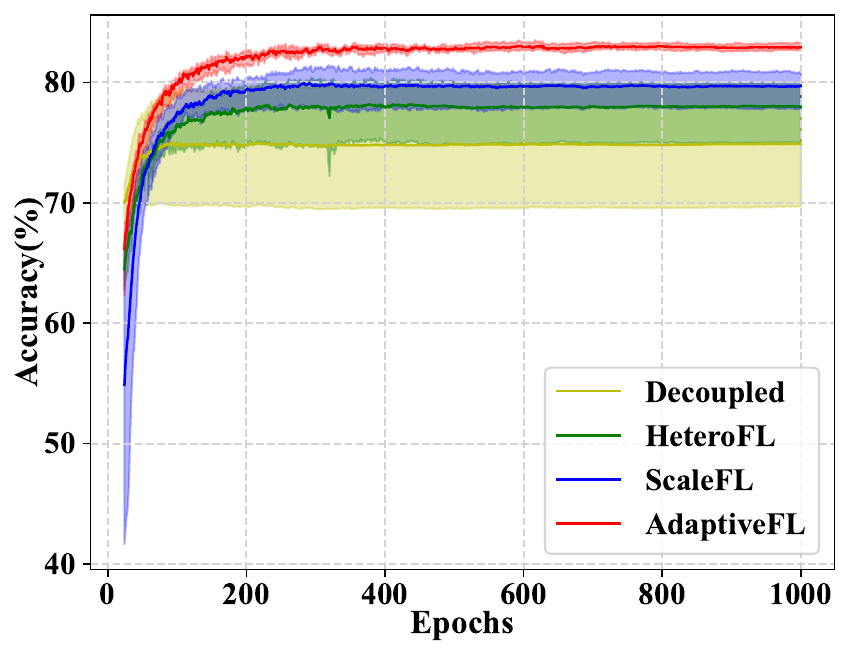}
    \vspace{-0.25in}
    \caption{CIFAR-10, IID}
    \label{fig1:vgg-cifar10-iid}
  \end{subfigure}%
  \hspace{0.1in}
  \begin{subfigure}{0.4\linewidth}
    \centering
    \includegraphics[width=\linewidth]{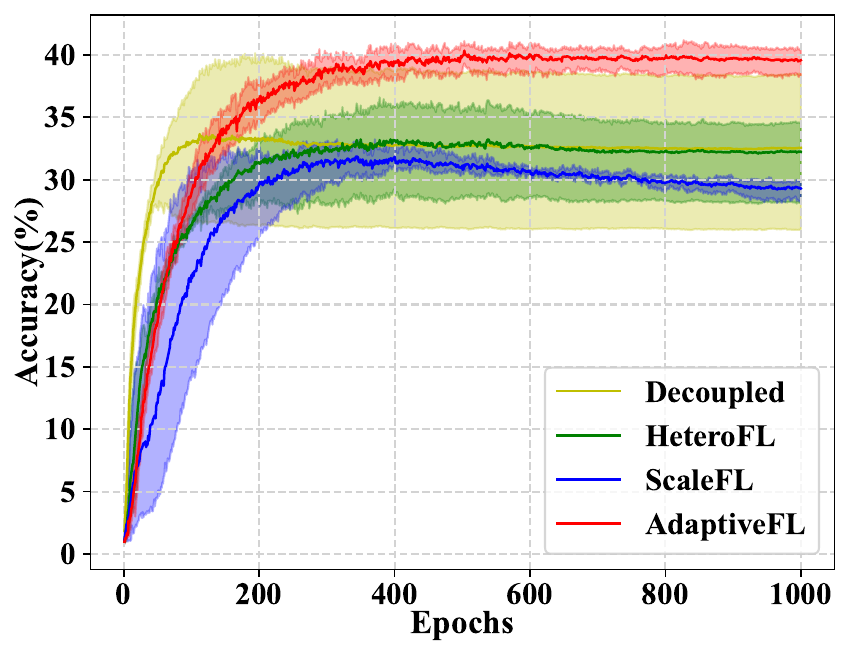}
    \vspace{-0.25in}
    \caption{CIFAR-100, IID}
    \label{fig2:vgg-cifar100-iid}
  \end{subfigure}


  \begin{subfigure}{0.4\linewidth}
    \centering
    \includegraphics[width=\linewidth]{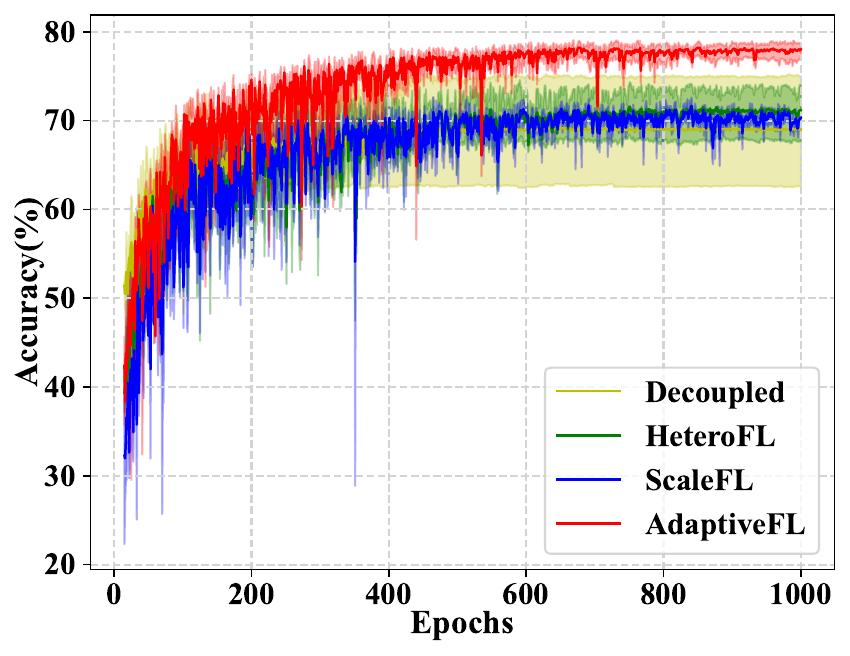}
    \vspace{-0.25in}
    \caption{CIFAR-10, $\alpha=0.3$}
    \label{fig3:vgg-cifar10-niid03}
  \end{subfigure}%
   \hspace{0.1in}
  \begin{subfigure}{0.4\linewidth}
    \centering
    \includegraphics[width=\linewidth]{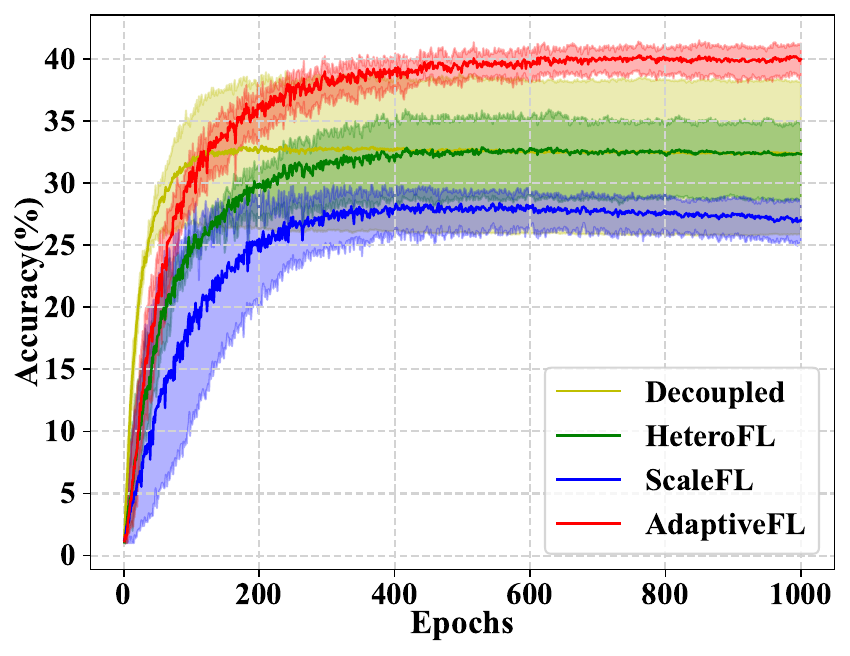}
    \vspace{-0.25in}
    \caption{CIFAR-100, $\alpha=0.3$}
    \label{fig4:vgg-cifar100-niid03}
  \end{subfigure}
  \vspace{-0.15in}
  \caption{Learning curves of AdaptiveFL and three baselines.}
  \label{fig: comparison for different methods}
\end{figure}

\begin{figure}[h]
  \centering
  \includegraphics[width=\linewidth]{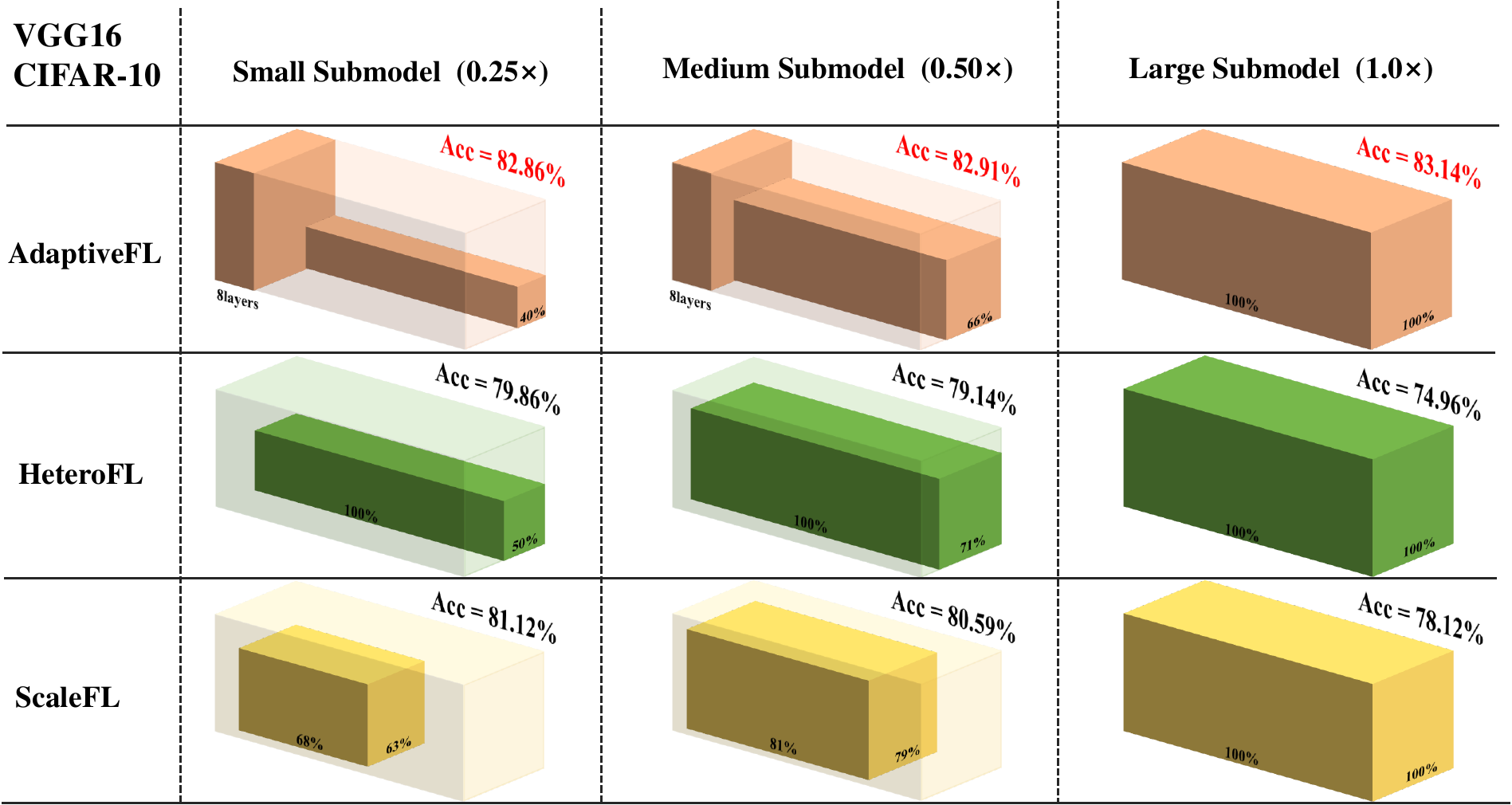} 
  \caption{Comparison of submodels at different levels.}
  \label{fig:Accuracies of local submodels}
\end{figure}

\textbf{Submodel Performance.}
Figure \ref{fig: comparison for different methods} shows
the learning trends of all methods on CIFAR-10/100 based on  VGG16, 
where solid lines represent the ``avg'' accuracy of submodels. 
We can find that AdaptiveFL can achieve the best inference performance with 
the least variations for both non-IID and IID scenarios. 
%
For different heterogeneous FL methods, 
Figure \ref{fig:Accuracies of local submodels} presents
the shapes of VGG16 submodels together with their test accuracy information. 
We can find that AdaptiveFL consistently outperforms other heterogeneous FL methods under the premise of satisfying the resource constraints, which indicates the effectiveness of our fine-grained width-wise model pruning mechanism. 
Interestingly, we can find that 
1.0$\times$ large models of HeteroFL and ScaleFL perform worse instead their 0.25$\times$ small counterparts.
Conversely, as the model size increases, AdaptiveFL can achieve better results, indicating that it can smoothly transfer the knowledge learned by the submodels into large models.

\subsection{Impacts of Different Configurations} \label{subsection: Impacts of Different Configurations}

\begin{figure}[h]  
  \centering
  \begin{subfigure}{0.4\linewidth}
    \centering
    \includegraphics[width=\linewidth]{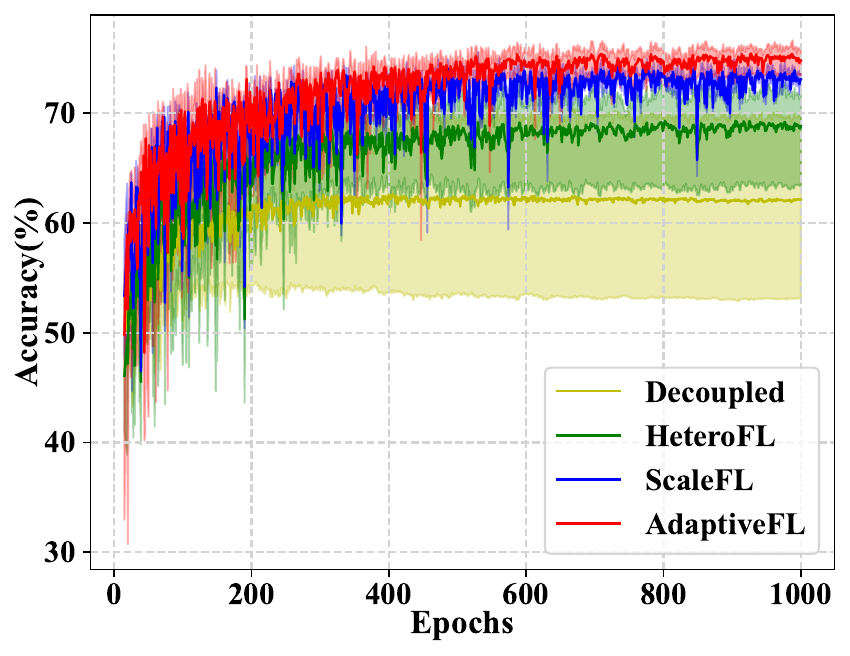}
    \vspace{-0.25in}
    \caption{$K$ = 50}
  \end{subfigure}%
  \hspace{0.1in}
  \begin{subfigure}{0.4\linewidth}
    \centering
    \includegraphics[width=\linewidth]{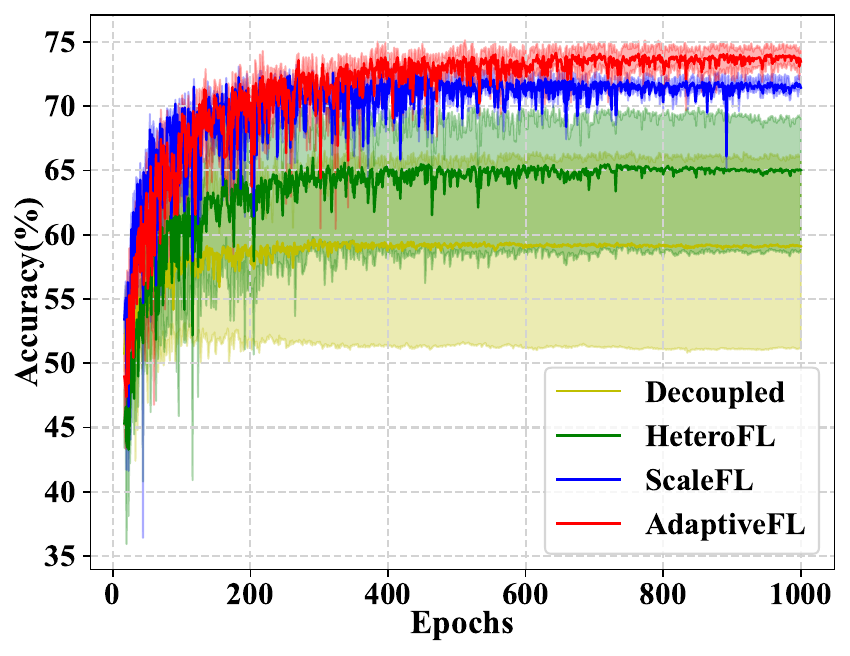}
    \vspace{-0.25in}
    \caption{$K$ = 100}
  \end{subfigure}
  \begin{subfigure}{0.4\linewidth}
    \centering
    \includegraphics[width=\linewidth]{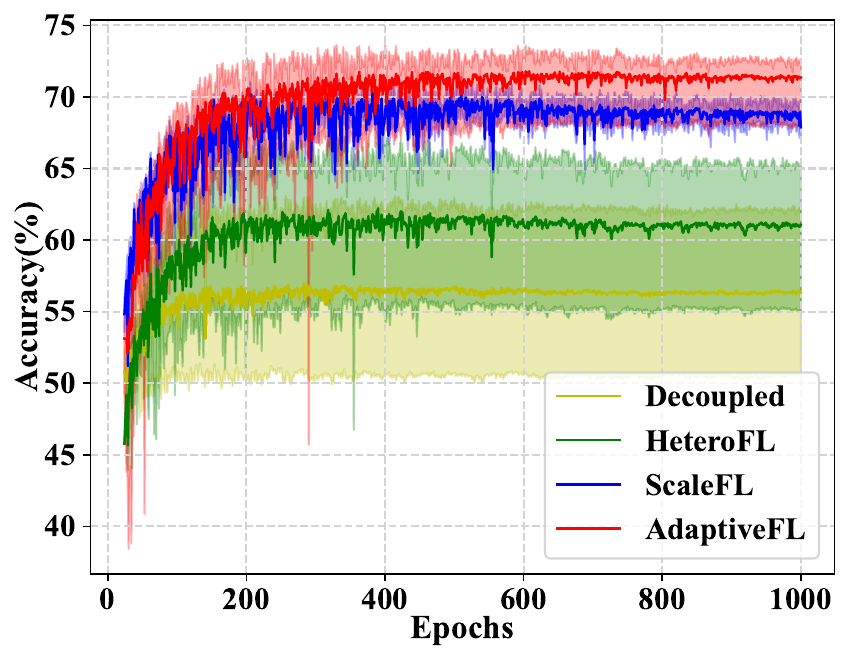}
    \vspace{-0.25in}
    \caption{$K$ = 200}
  \end{subfigure}%
  \hspace{0.1in}
  \begin{subfigure}{0.4\linewidth}
    \centering
    \includegraphics[width=\linewidth]{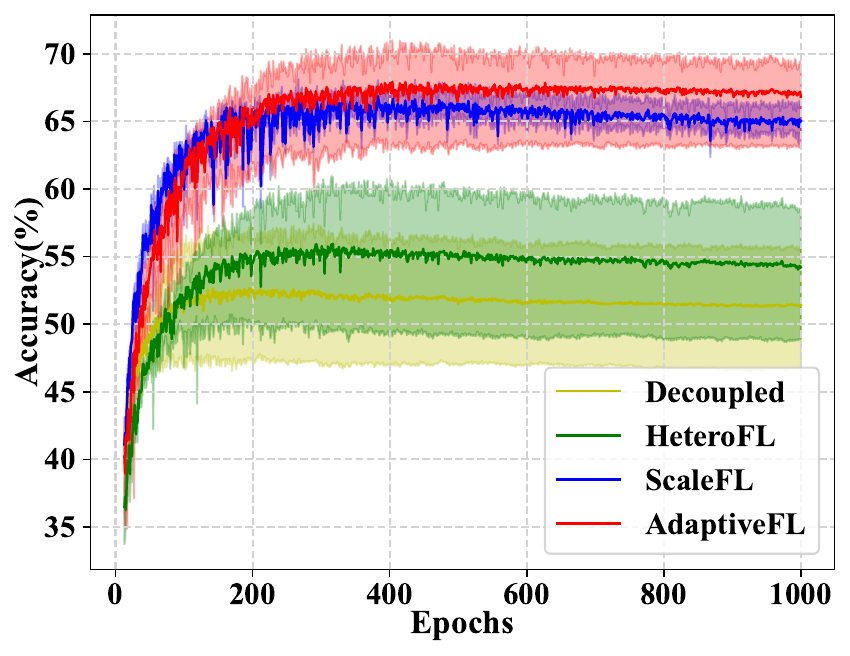}
    \vspace{-0.25in}
    \caption{$K$ = 500}
  \end{subfigure}
    \vspace{-0.1in}
  \caption{Learning curves on different numbers of involved devices.}
  \label{fig: Learning curves for different numbers of simultaneously training clients}
  \vspace{-0.1in}
\end{figure}

\textbf{Numbers of Participating Clients.} 
To evaluate the scalability of AdaptiveFL, we conducted experiments considering different numbers of participating clients, i.e., $K$ = 50, 100, 200, and 500, respectively, on CIFAR-10 using ResNet18.
Figure \ref{fig: Learning curves for different numbers of simultaneously training clients} compares AdaptiveFL with three baselines within a non-IID scenario ($\alpha=0.6$), where  
%
 AdaptiveFL can always achieve the highest accuracy.

\textbf{Proportions of Different Devices.} 
Table \ref{Influence of different proportions of device types on the Cifar-10 dataset} shows the performance of  AdaptiveFL with different proportions (i.e., 8:1:1, 1:8:1, 1:1:8, and 4:3:3) of weak, medium, and strong devices on CIFAR-10. We can find that AdaptiveFL can achieve the best test accuracy in all cases. 
Note that as the proportion of strong devices increases, the global model performance of all FL methods improves.

\begin{table}[h]
\caption{Performance comparison (\%) under different proportions.}
\label{Influence of different proportions of device types on the Cifar-10 dataset}
\scriptsize
\begin{tabular}{l|cccccccc}
\hline
\multicolumn{1}{c|}{\multirow{3}{*}{Algorithm}} & \multicolumn{8}{c}{Proportion} \\ \cline{2-9} 
\multicolumn{1}{c|}{} & \multicolumn{2}{c|}{\textbf{4:3:3}} & \multicolumn{2}{c|}{\textbf{8:1:1}} & \multicolumn{2}{c|}{\textbf{1:8:1}} & \multicolumn{2}{c}{\textbf{1:1:8}} \\ \cline{2-9} 
\multicolumn{1}{c|}{} & avg & \multicolumn{1}{c|}{full} & avg & \multicolumn{1}{c|}{full} & avg & \multicolumn{1}{c|}{full} & avg & full \\ \hline
All-Large & - & \multicolumn{1}{c|}{79.76} & - & \multicolumn{1}{c|}{79.76} & - & \multicolumn{1}{c|}{79.76} & - & 79.76 \\
HeteroFL & 77.98 & \multicolumn{1}{c|}{74.96} & 72.43 & \multicolumn{1}{c|}{64.44} & 75.94 & \multicolumn{1}{c|}{65.96} & 81.26 & 81.12 \\
ScaleFL & 79.94 & \multicolumn{1}{c|}{78.12} & 75.89 & \multicolumn{1}{c|}{72.03} & 78.40 & \multicolumn{1}{c|}{72.30} & 82.55 & 82.81 \\
\textbf{AdaptiveFL} &\textbf{ 82.95} & \multicolumn{1}{c|}{\textbf{83.14}} & \textbf{81.62} & \multicolumn{1}{c|}{\textbf{81.93}} & \textbf{82.78} & \multicolumn{1}{c|}{\textbf{82.89}} & \textbf{82.82} & \textbf{83.24} \\ \hline
\end{tabular}
\end{table}

\subsection{Ablation Study} \label{subsection: Ablation study}

\textbf{Ablation of Fine-grained Pruning.} 
To evaluate both fine- and coarse-grained pruning, we set $p$ to 3 and 1 for each level, respectively.
Table \ref{table: Comparison on AdaptiveFL with fine-grained and coarse-grained segmentation} presents the ablation results of AdaptiveFL considering the effect of our fine-grained pruning method, showing that 
the fine-grained pruning method can achieve up to 9.38\% inference accuracy improvements for AdaptiveFL. Note that the fine-grained methods can consistently achieve 
better inference results than their coarse-grained counterparts, since  
the fine-grained ones can better transfer the knowledge of small models to large models.


\begin{table}[h]
\vspace{-0.05in}
\caption{Ablation of fine-grained pruning (accuracy on ``full'').}
\vspace{-0.1in}
\label{table: Comparison on AdaptiveFL with fine-grained and coarse-grained segmentation}
\scriptsize
\addtolength{\tabcolsep}{1pt}
\begin{tabular}{c|cc|ccc}
\hline
\multirow{2}{*}{Dataset}   & \multirow{2}{*}{Model}    & \multirow{2}{*}{Grained} & \multicolumn{3}{c}{Distribution}      \\ \cline{4-6} 
                           &                           &                            & IID          &$\alpha$ = 0.6 &$\alpha$ = 0.3      \\ \hline
\multirow{4}{*}{CIFAR-10}  & \multirow{2}{*}{VGG16}    & coarse               & 80.1         & 78.9    & 74.27        \\
                           &                           & fine              & 83.14 (\textbf{+3.04}) & 81.31 (\textbf{+2.41})   & 78.99 (\textbf{+4.72}) \\ \cline{2-6} 
                           & \multirow{2}{*}{ResNet18} & coarse              & 72.43        & 71.92   & 66.07        \\
                           &                           & fine               & 77.2 (\textbf{+4.77})  & 74.89 (\textbf{+2.97})   & 70.97 (\textbf{+4.9})  \\ \hline
\multirow{4}{*}{CIFAR-100} & \multirow{2}{*}{VGG16}    & coarse              & 38.91         & 39.43    & 39.29        \\
                           &                           & fine               & 40.93 (\textbf{+2.02}) & 38.88 (\textbf{-0.55})   & 41.17 (\textbf{+1.88}) \\ \cline{2-6} 
                           & \multirow{2}{*}{ResNet18} & coarse               & 31.77        & 35.52   & 34.73        \\
                           &                           & fine              & 41.15 (\textbf{+9.38})  & 39.56 (\textbf{+4.04})   &39.65 (\textbf{+4.92})  \\ \hline
\end{tabular}
\vspace{-0.05in}
\end{table}

\textbf{Ablation of RL-based Client Selection.}
To evaluate the effectiveness of our RL-based client selection strategy, we developed four variants of AdaptiveFL: i) ``AdaptiveFL+Greedy'' that always dispatches the largest model for each selected client; ii) “AdaptiveFL+Random” that selects clients for local training randomly; iii) “AdaptiveFL+C” that selects clients only based on curiosity reward; and iv) “AdaptiveFL+S” that selects clients only using resource rewards.
Moreover, we use “AdaptiveFL+CS” to indicate the original AdaptiveFL implemented in Algorithm \ref{algorithm: AdaptiveFL}.

\begin{figure}[h]  
  \centering
  \begin{subfigure}{0.45\linewidth}
    \centering
    \includegraphics[width=0.99\linewidth]{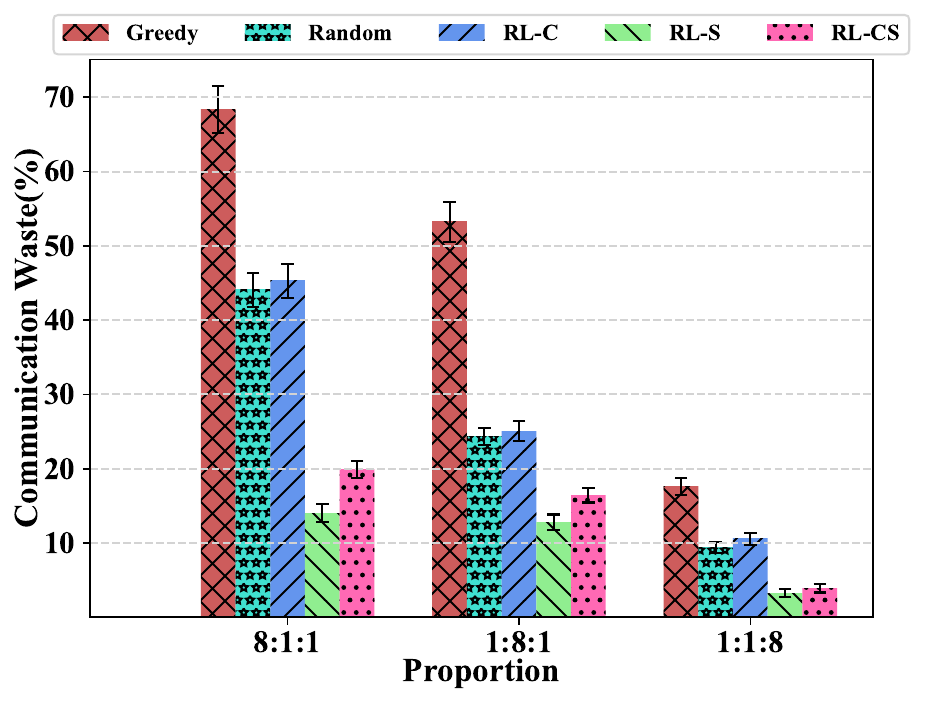}
    \vspace{-0.2in}
    \caption{Communication waste}
    \label{ablation_resource_waste}
  \end{subfigure}
    \hspace{0.1in}
  \begin{subfigure}{0.45\linewidth}
    \centering
    \includegraphics[width=0.99\linewidth]{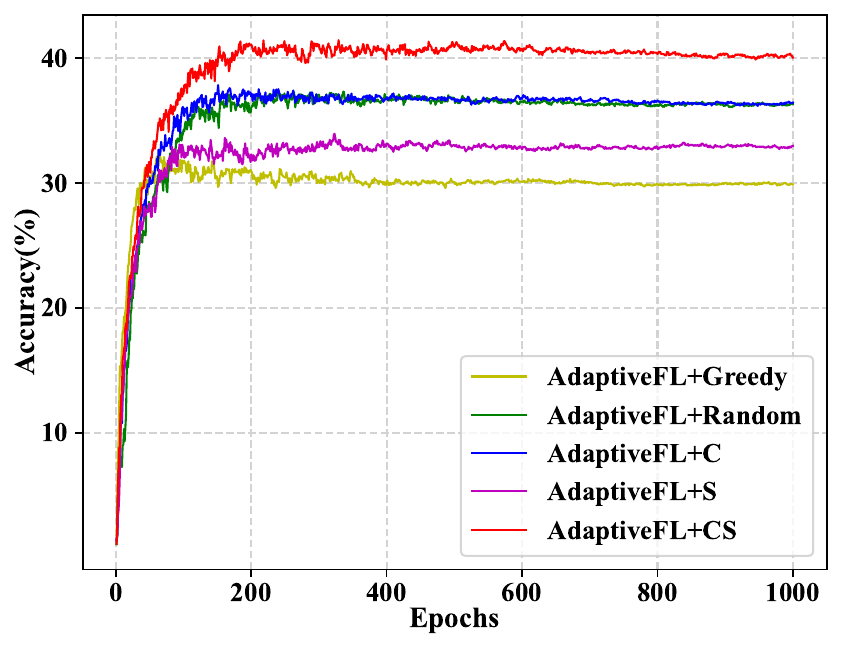}
    \vspace{-0.2in}
    \caption{Selection strategy}
    \label{ablation_RL_client_selection}
  \end{subfigure}
\vspace{-0.15in}
    \caption{Ablation study for RL-based client selection.}
  \label{fig: Comparison with different strategy}
  \vspace{-0.1in}
\end{figure}

Figure \ref{fig: Comparison with different strategy} presents the ablation study results on CIFAR-100 with ResNet18 following IID distribution. 
To indicate  the similarity between a sending model and its corresponding
receiving model, we introduce 
a new metric called \textit{communication waste rate}, defined as 
 ``$1-\sum(\operatorname{size}(\text{$ML_{back}$}))/\sum(\operatorname{size}(\text{$ML_{send}$}))$''. The lower the rate, the closer the two models are, leading to less local pruning efforts. From Figure \ref{fig: Comparison with different strategy}, we can find that our approach can achieve the highest accuracy with low communication waste (second only to RL-S).


\subsection{Evaluation on Real Test-bed}
Based on our real test-bed platform, 
we conducted experiments on a non-IID  IoT dataset (i.e.,  Widar~\cite{fedaiot}) with MobileNetV2~\cite{mobilenetv2} models. We assumed that the FL-based  AIoT system has 
17 devices, each training round involves 10 selected devices, whose   
detail heterogeneous configurations are shown in Table \ref{table: Composition of real test-bed platform}.

\begin{table}[h]
\caption{Real test-bed platform configuration.}
\label{table: Composition of real test-bed platform}
\vspace{-0.1in}
\footnotesize
\addtolength{\tabcolsep}{-1pt}
\begin{tabular}{c|c|cl|c|c}
\hline
\multicolumn{1}{c|}{\textbf{Type}} & \multicolumn{1}{c|}{\textbf{Device}} & \multicolumn{2}{c|}{\textbf{Comp}}                 & \textbf{Mem} & \textbf{Num} \\ \hline
Client-Weak   & Raspberry Pi 4B   & \multicolumn{2}{c|}{ARM Cortex-A72 CPU}   & 2G  & 4   \\
Client-Medium & Jetson Nano       & \multicolumn{2}{c|}{128-core Maxwell GPU} & 8G  & 10  \\
Client-Strong & Jetson Xavier AGX & \multicolumn{2}{c|}{512-core NVIDIA GPU}  & 32G & 3   \\ \hline
Server        & Workstation       & \multicolumn{2}{c|}{NVIDIA RTX 4090 GPU}  & 64G & 1   \\ \hline
\end{tabular}
\end{table}

\begin{figure}[h]  
\vspace{-0.1in}
  \centering
  \begin{subfigure}{0.45\linewidth}
    \centering
    \includegraphics[width=\linewidth]{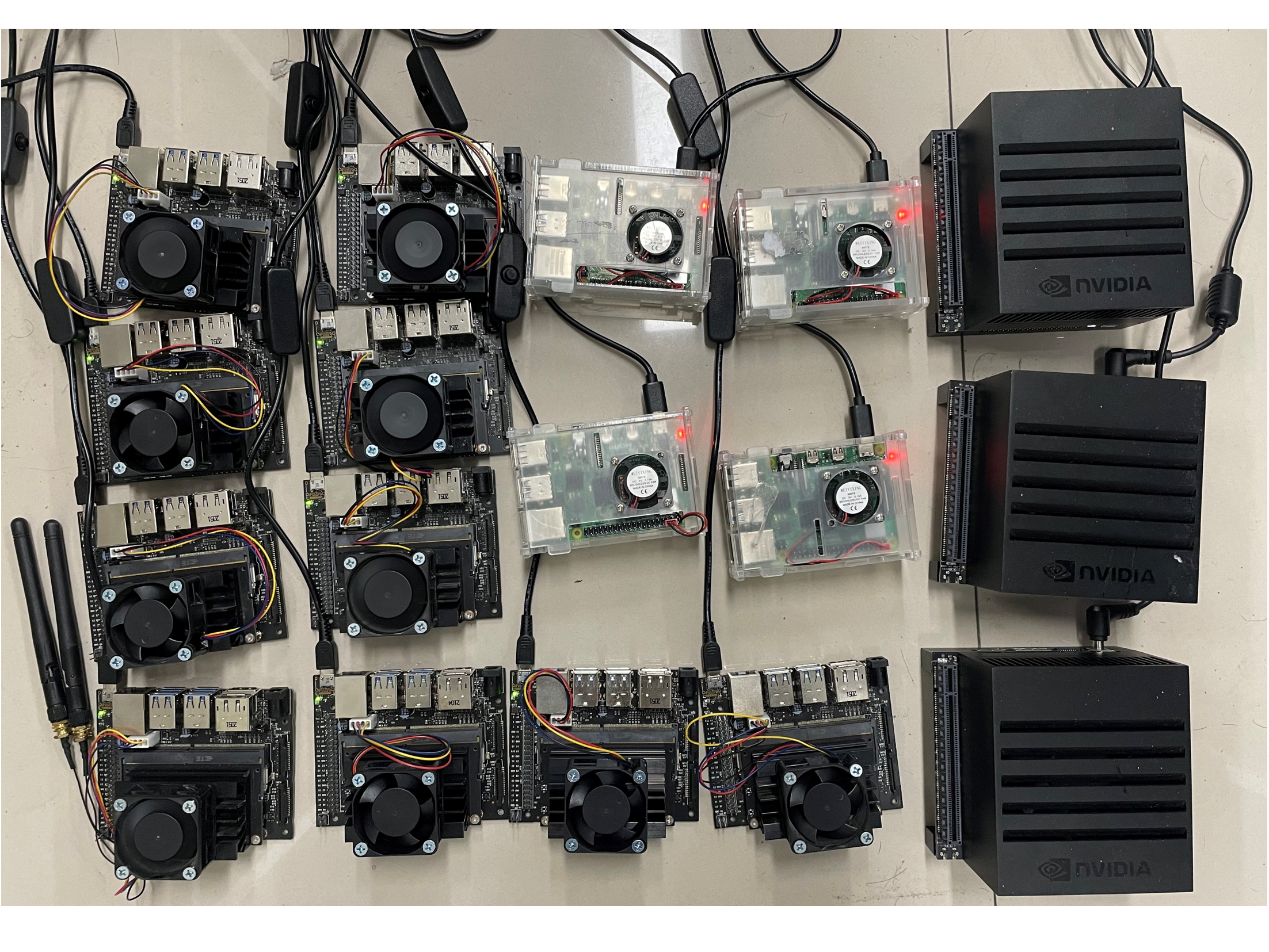}
    \vspace{-0.2in}
    \caption{Real test-bed platform}
    \label{Real test-bed platform}
  \end{subfigure}%
     \hspace{0.1in}
  \begin{subfigure}{0.45\linewidth}
    \centering
    \includegraphics[width=\linewidth]{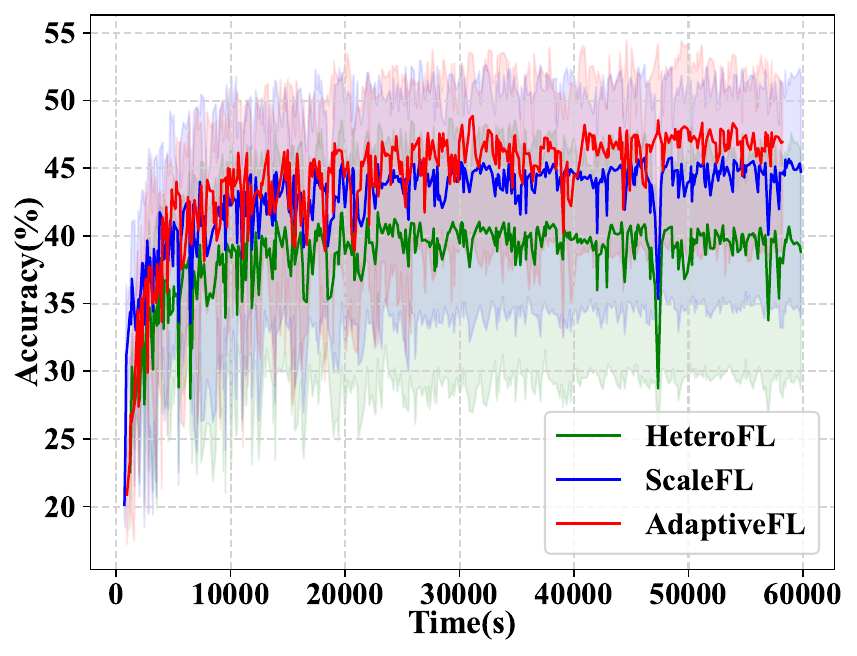}
    \vspace{-0.2in}
    \caption{Learning curves}
    \label{ablation_RL_client_selection}
  \end{subfigure}
  \vspace{-0.15in}
    \caption{Real test-bed experiment.}
  \label{fig: Real test-bed experiment}
  \vspace{-0.2in}
\end{figure}

Figure ~\ref{fig: Real test-bed experiment} presents the AIoT devices used in our experiment and the comparison results obtained from our real test-bed platform. 
We can observe that AdaptiveFL achieves the best inference results even in real scenarios compared to all baselines.


\section{Conclusion}
This paper presented a novel Federated Learning (FL) approach named AdaptiveFL to enable effective knowledge sharing among heterogeneous devices for large-scale Artificial Intelligence of Things (AIoT) applications, considering the varying on-the-fly hardware resources of AIoT devices. 
Based on our proposed fine-grained width-wise model pruning mechanism, 
AdaptiveFL supports the generation of 
different local models, which will be selectively dispatched to
their AIoT device counterparts in an adaptive manner according to their available local training resources. Experimental results show that our approach can achieve better inference performance than state-of-the-art heterogeneous FL methods.

\section*{Acknowledgment}
This research is supported by the Natural Science Foundation of China (62272170), ``Digital Silk Road'' Shanghai International Joint Lab of Trustworthy Intelligent Software (22510750100), and
the National Research Foundation Singapore and DSO National Laboratories under the AI Singapore Programme (AISG Award No: AISG2-RP-2020-019).
Ming Hu and Mingsong Chen are the corresponding authors.

\bibliographystyle{unsrt}
\bibliography{references_old}

\end{document}